\documentclass[letterpaper, 10 pt, conference]{ieeeconf}  

\pdfoutput=1

\IEEEoverridecommandlockouts                              

\usepackage{graphics} 
\usepackage{epsfig} 
\usepackage{amsmath} 
\usepackage{amssymb}  
\usepackage{cite}
\usepackage[english]{babel}
\usepackage{graphicx}
\usepackage{hyperref}       
\usepackage{url}            
\usepackage{nicefrac}       
\usepackage{microtype}      
\usepackage{dsfont}
\usepackage{algorithm}
\usepackage{algorithmic}

\title{\LARGE \bf
Sim-to-Real Transfer Learning using Robustified Controllers in Robotic Tasks involving Complex Dynamics
}

\author{Jeroen van Baar, Alan Sullivan, Radu Cordorel, Devesh Jha, Diego Romeres and Daniel Nikovski
\thanks{Mitsubishi Electric Research Laboratories (MERL), Cambridge, MA 02139, USA
        {\tt\small jeroen@merl.com}}
}

\begin{document}

\maketitle
\thispagestyle{empty}
\pagestyle{empty}

\begin{abstract}
  Learning robot tasks or controllers using deep reinforcement learning has been proven effective in simulations. Learning in simulation has several advantages. For example, one can fully control the simulated environment, including halting motions while performing computations. Another advantage when robots are involved, is that the amount of time a robot is occupied learning a task---rather than being productive---can be reduced by transferring the learned task to the real robot. Transfer learning requires some amount of fine-tuning on the real robot. For tasks which involve complex (non-linear) dynamics, the fine-tuning itself may take a substantial amount of time. In order to reduce the amount of fine-tuning we propose to learn robustified controllers in simulation. Robustified controllers are learned by exploiting the ability to change simulation parameters (both appearance and dynamics) for successive training episodes. An additional benefit for this approach is that it alleviates the precise determination of physics parameters for the simulator, which is a non-trivial task. We demonstrate our proposed approach on a real setup in which a robot aims to solve a maze game, which involves complex dynamics due to static friction and potentially large accelerations. We show that the amount of fine-tuning in transfer learning for a robustified controller is substantially reduced compared to a non-robustified controller.
\end{abstract}

\section{Introduction}

Teaching robots to perform challenging tasks has been an active topic of research. In particular, it has recently been demonstrated that reinforcement learning (RL) coupled with deep neural networks is able to learn policies (controllers) which can successfully perform tasks such as pick and fetch.

Robots may be slow, dangerous, can damage themselves and they are expensive. When a robot is learning a task, it needs to be taken out of production. Learning policies using model-free deep RL typically requires many samples to explore the sequential decision making space. Model-free RL applied to tasks that involve complex dynamics, require even more samples to learn adequate policies compared to tasks involving (largely) linear dynamics. Directly learning on robots may thus be very costly.

In order to reduce the time required for learning on a real robot, training can be performed in simulation environments. The learned policy is then transferred to the real world domain. Modern graphics cards and sophisticated physics engines enable the simulation of complex tasks. Learning with simulators has several advantages. The rendering and physics engines are capable of computing simulations faster than real-time. This helps to reduce overall training times. Recent deep reinforcement learning algorithms allow agents to learn in parallel~\cite{mnih:2016:A3C}, which reduces training times. Furthermore, both appearance and physics can be controlled in simulation. For example the lighting condition, or the friction of an object can be changed, or the entire simulation can be halted to allow for computation of updates.

Appearance, complex dynamics, and robot motor movements in the real world can only be simulated up to some approximation. Simulation to real world transfer thus requires fine-tuning on real data. Furthermore, real setups involving various components, experience delays which are hard to determine exactly. For example, the delay introduced by the acquisition system, where some time has passed before the acquired image is available for processing by the algorithm.

By randomization of the appearance, physics and system parameters during reinforcement learning on simulation data, \textit{robustified} policies can be learned. This is analogous to training a deep convolutional neural network to classify objects regardless of the background in the input images. We found that robustified policies can greatly reduce the amount of time for fine-tuning in transfer learning. Reducing the fine-tuning time in transfer learning becomes especially important for tasks involving complex dynamics.

\begin{figure*}
  \centering
  \includegraphics[width=\textwidth]{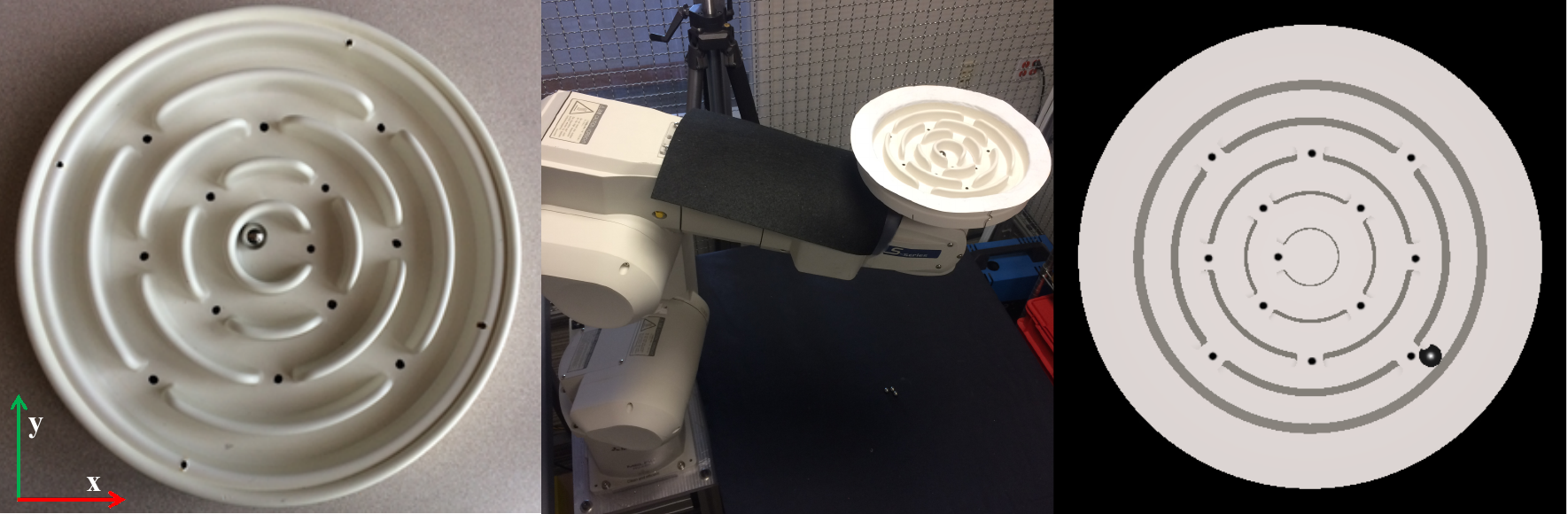}
  \caption{Marble maze game. (\textbf{Left}) Top view of the marble maze after a plexiglass top has been removed (leaving holes in the outermost edge). A paper rim is used to cover the holes. The black dots in each gate between rings are used for alignment. The view also shows the world aligned \(x\) and \(y\) axes. (\textbf{Middle}) The marble maze mounted on the robot arm. (\textbf{Right}) A rendering of the simulated marble maze under some chosen lighting conditions (without added noise).}\label{fig:maze}
\end{figure*}

We demonstrate our proposed approach on a challenging task of a robot learning to solve a marble maze game. The maze game is shown in Figure~\ref{fig:maze}. The marbles are subject to static and rolling friction, acceleration, and collisions (with other marbles and with the maze geometry). A simulator simulates the physics of the marbles in the maze game, and renders the results to images. We learn to solve the game from scratch using deep reinforcement learning. A modified version of the deep reinforcement learning is used to learn directly on real robot hardware. We learn both a robustified and non-robustified policy in simulation and compare the times required for fine-tuning after transferring the policy to the real world.

In the remainder of this paper we will refer to learning on simulated data / environments as \textit{offline} learning, and learning on real data / environments will be referred to as \textit{online} learning. Transfer learning (TL) with fine-tuning on real data therefore constitutes both offline as well as online learning.

\section{Related Work}

Our work is inspired by the recent advances in deep reinforcement learning, learning complicated tasks and achieving (beyond) human level performance on a variety of tasks~\cite{mnih:2015:humanlevel, mnih:2016:A3C, silver:2016:mastergo, silver:2017:masteringgonohuman}.

TL has been an active area of research in the context of deep learning. For example, tasks such as object detection and classification can avoid costly training time by using pre-trained networks and fine-tuning~\cite{razavian:2014:cnnofftheshelf, yosinski:2014:howtransferable}, where typically only the weights in the last couple of layers are updated. TL from simulated to real has also been applied to learn robot tasks~\cite{rusu:2016:sim2realpronets, zhang:2017:sim2realclutter, zhang:2016:modulardeepsim2real, bousmalis:2017:domainlearn}. To reduce the time required for fine-tuning in TL, the authors in~\cite{shrivastava:2016:simadversarial} propose to make simulated data look more like the real world.

In~\cite{christiano:2016:deepinversedynamics} the authors acknowledge that training robot tasks on simulated data alone does not readily transfer to the real world. They propose a form of fine-tuning where the inverse dynamics for the real robot are recovered. It requires a simulator and training which produces reasonable estimates of the real world situation. The drawback of this method is that it requires long online training times, whereas our goal is to minimize the duration of the online training time.

By randomization of the appearance, the learning can become robust against appearance changes and readily transfer to the real world domain~\cite{james:2017:sim2realtransfer, sadeghi:2016:flightworealimage}. The method proposed in~\cite{rajeswaran:2016:epopt} exploits an ensemble of simulated source domains and adversarial training to obtain robust policies. This policy search approach relies on trajectories and roll-outs which solve the task. The approach proposed in~\cite{lowrey:2018:rlmanupilation} uses model-based RL to learn a controller entirely in simulation, allowing for zero-shot TL. Since we are considering tasks involving (much) more complex dynamics, we instead follow a similar approach as~\cite{peng:2018:dynrand}, and perform randomization of appearance, physics and system parameters with model-free RL.

Model-agnostic meta-learning (MAML)~\cite{finn:2017:maml}, aims to learn a meta-policy that can be quickly adapted to new (but similar) tasks. In the case of complex dynamics it is not clear how easily MAML could be applied. Appearance and dynamics randomization can be considered as forms of meta-learning. Other approaches aim to learn new tasks, or refine previously learned tasks, without "forgetting", e.g.,~\cite{li:2016:learnwoforget}. Our emphasis instead is on reducing the amount of time required for fine-tuning in TL.

Our simulator provides observations of the state in simulation, similar to the real world. In~\cite{pinto:2017:asymactorcritic} the critic receives full states, whereas the actor receives observations of states. Coupled with appearance randomization, zero-shot transfer can be achieved. The full state requires that the physics parameters to produce complex dynamics match those of the real world. However, precisely determining the physics parameters is non-trivial.

Formulating reward functions is not straightforward. The authors in ~\cite{fu:2017:robustrewards} propose to discover robust rewards to enable the learning of complicated tasks. Adding additional goals (sub-goals), basically a form of curriculum learning~\cite{bengio:2009:CL}, can improve the learning as well~\cite{andry:2017:HER}. The latter approach may be applied to break up the goal of a marble maze into stages. However, in this paper we show that a simple reward function which governs the overall goal of the game is sufficient.

The authors in~\cite{brockman:2016:openaigym} propose a game-like environment for generating synthetic data for benchmark problems related to reinforcement learning. We developed our simulator along the same lines as~\cite{brockman:2016:openaigym}.

In~\cite{jharomeres:2018:modelbasedmaze} the authors propose to model both the dynamics and control in order to solve the marble maze game. This is a complementary approach to the TL approach proposed in this paper, and we believe that each approach has its own strengths and weaknesses.

\section{Preliminaries}

We briefly review some concepts from (deep) reinforcement learning (RL) using model-free asynchronous actor-critic, and define some terminology that we will use in the remainder of this paper. In the next section we will discuss our approach.

\subsection{Reinforcement Learning}

In RL an agent interacts with an environment, represented by a set of states \(\mathcal{S}\), taking actions from an action set \(\mathcal{A}\), and receiving rewards \(r: \mathcal{S} \times \mathcal{A} \rightarrow \mathbb{R}\). The environment is governed by (unknown) state transition probabilities \(p(s'|s, a)\). The agent aims to learn a (stochastic) policy \(\pi(a|s)\), which predicts (a distribution over) actions \(a\) based on state \(s\). The goal for the agent is to learn a policy which maximizes the expected return \(\mathds{E}[R_t]\), where the return \(R_t = \sum_{k=0}^{\infty} \gamma^k r_{t+k}\) denotes the discounted sum of future rewards, with discount factor \(\gamma\).

To determine for a given policy \(\pi\) \textit{how good it is} to be in a certain state, or \textit{how good it is} to take a certain action in a certain state, RL depends on two value functions: a state-value function \(V^{\pi}(s) = \mathds{E}_{\pi}[\sum_{k=0}^{\infty}\gamma^k R_{t+k+1}|S_t=s]\) and an action-value function \(Q^{\pi}(s, a) = \mathds{E}_{\pi} [\sum_{k=0}^{\infty} \gamma^k R_{t+k+1}|S_t=s, A_t=a]\). For Markov decision processes, the value functions can be written as a recursion of expected rewards, e.g., \(V^{\pi}(s) = R(s, \pi(s)) + \gamma \sum_{s'} p(s'|s, \pi(s)) V^{\pi}(s')\), where \(s\) denotes the current state, and \(s'\) denotes the next state. The recursive formulations are Bellman equations. Solving the Bellman optimality equations would give rise to the optimal policy \(\pi^*\). For details we refer the reader to~\cite{sutton:1998:IRL}

We consider the case where agents interact with the environment in episodes of finite length. The end of an episode is reached if the agent arrives at the timestep of maximum episode length, or the goal (terminal state) is achieved. In either case, the agent restarts from a new initial state.

\subsection{Deep RL using Advantage Actor-Critic}

In \cite{mnih:2016:A3C} the authors propose the asynchronous advantage actor-critic algorithm. The algorithm defines two networks: a policy network \(\pi(a|s, \theta_p)\) with network parameters \(\theta_p\), and a value network \(V(s|\theta_v)\) with network parameters \(\theta_v\). This policy-based model-free method determines a reduced variance estimate of \(\nabla_{\theta_p} \mathds{E}[R_t]\) as \(\nabla_{\theta_p}\log \pi(a_t|s_t, \theta_p)(R_t - b_t(s_t))\)~\cite{williams:1992:ssg}. The return \(R_t\) is an estimate of \(Q^{\pi}\) and the \textit{baseline} \(b_t\) is a learned estimate of the value function \(V^{\pi}\). The policy \(\pi\) is referred to as the \textit{actor}, and value function estimate \(V^{\pi}\) as the \textit{critic}.

The authors in \cite{mnih:2016:A3C} describe an algorithm where multiple agents learn in parallel, and each agent maintains local copies of the policy and value networks. Agents are trained on episodes of maximum length \(L_{e}\). Within each episode, trajectories are acquired as sequences \(\tau = (s_1, a_1, r_1, s_2, a_2, r_2, \dots, s_{L_{se}}, a_{L_{se}}, r_{L_{se}})\), of maximum length \(L_{se}\). Rather than the actual state, the inputs are observations (images) of the state, and a forward pass of each image through the agent's local policy network results in a distribution over the actions. Every \(L_{se}\) steps, the parameters of the global policy and value networks are updated and the agent synchronizes its local copy with the parameters of the global networks. The current episode ends after \(L_{e}\) steps, or when the terminal state is reached, and then a new episode starts. This episodal learning is repeated until the task is solved consistently. See \cite{mnih:2016:A3C} for further details.

\section{Deep Reinforcement Learning for a Task with Complex Dynamics}

\subsection{Setting up the Task}
\label{subsec:tasksetup}

The task we aim to learn is to solve a marble maze game, see Figure~\ref{fig:maze}. Solving the game means that the marble(s) are maneuvered from the outermost ring, through a sequence of gates, into the center. Due to static and dynamic friction, acceleration, damping, and the discontinuous geometry of the maze, the dynamics are (highly) complex and difficult to model. To solve the marble maze game using model-free RL we can define a reward function as:

\begin{equation}\label{reward-func}
  r =
  \begin{cases}
    -1, & \text{if through gate away from the goal} \\
    +1, & \text{if through gate towards the goal} \\
    0, & \text{otherwise}
  \end{cases}
\end{equation}

This sparse reward function is general and does not encode any information about the actual geometry of the game. The action space is discretized into five actions. The first four actions constitute \(1^{\circ}\) rotation increments, clockwise and counterclockwise around the \(x\), and \(y\) axes up to a fixed maximum angle. Figure~\ref{fig:maze}--Left shows the orientation of the \(x\), and \(y\) axes with respect to the maze. The \(1^{\circ}\) increment is sufficient to overcome the static friction, while simultaneously avoiding accelerations that are too large. We define a fifth action as \textit{no-op}, i.e., maintain the current orientation of the maze. We empirically determined the fixed maximum angle to be \(5^{\circ}\) in either direction.

\subsection{Deep Reinforcement Learning on Simulated Robot Environments}

In order to learn a robustified policy in simulation, we adopt the idea of randomization from~\cite{peng:2018:dynrand, james:2017:sim2realtransfer, sadeghi:2016:flightworealimage}. We implemented two learning schemes. In the first scheme, each agent was assigned different parameters which were kept fixed for the duration of learning. In the second scheme, the physics and appearance parameters are randomly sampled from a pre-determined range, according to a uniform distribution, for each episode and each agent. We found that the second scheme produced robustified policies which adapted more quickly during fine-tuning on the real robot after transfer.

We use the asynchronous advantage actor-critic (\textsc{A3C}) algorithm to learn a policy for the marble maze game. To successfully apply reinforcement learning with sparse rewards, a framework of auxiliary tasks may be incorporated~\cite{jaderberg:2016:unreal}. One could consider path following as an auxiliary (dense reward) task. However, we aim to keep our approach as general as possible, and not rely on the geometry of the maze. We instead incorporate pixel change and reward prediction, as proposed by~\cite{jaderberg:2016:unreal}. Pixel change promotes taking actions which result in maximal change between images of consecutive states. In the context of the maze game, we aim to avoid selecting consecutive actions that would result in little to no marble motions. In addition, reward prediction aims to over-represent rewarding events to offset the sparse reward signal provided by the reward function. To stabilize learning and avoid settling into sub-optimal policies we employ the generalized advantage estimation as proposed by~\cite{schulman:2015:gae} together with entropy regularization with respect to the policy parameters~\cite{mnih:2016:A3C}.

\subsubsection{Robustified Policies}

At the start of each episode, for each agent, the parameter values for static friction, dynamic friction, damping and marble(s) mass are uniformly sampled from a range of values. We emulated a camera delay by rendering frames into a buffer. The camera delay was varied per episode and agent. During each episode the parameters are held constant. Each observation received from the simulator is corrupted by AGWN. We experimented with additional appearance changes, such as different light colors and intensities. We found that those changes had little effect on improving the time required for fine-tuning for our current setup.

\subsection{Deep Reinforcement Learning on Real Robot Environments}

\textsc{A3C} is an on-policy method, since the current policy \(\pi(s; \theta)\) is used in roll-outs (using an \(\epsilon\)-greedy exploration strategy) to obtain the current trajectory of length \(L_{se}\). For each update, \textsc{A3C} accumulates the losses for the policy and value networks over the trajectory and performs backpropagation of the losses to update the policy and value network parameters. The simulation is halted until the network parameters have been updated, and then roll-outs for the next trajectory continue using the updated policy \(\pi(s; \theta')\).

For a real robot setup we need to be able to compute an update, while simultaneously collecting the next trajectory, since we cannot halt the motion of the marble(s) during an update. We therefore adopt an off-policy approach for the real robot setups (see Algorithm~\ref{alg:offpolicy}).

\begin{algorithm}
\caption{Algorithm for off-policy \textsc{A3C}}
\label{alg:offpolicy}
\begin{algorithmic}
\STATE{\(\pi(s|\theta)\)---initialized or robustly learned in simulation}
\STATE{\(t \leftarrow 0\)}
\STATE{obtain \(\tau_t\) using \(\pi(s|\theta)\)}
\REPEAT
\WHILE[concurrently]{\(|\tau_{t+1}|<L_{se}\)}
\STATE{ compute update from \(\tau_t \rightarrow \pi(s|\theta')\)}
\STATE{ obtain \(\tau_{t+1}\) using \(\pi(s|\theta)\)}
\ENDWHILE
\STATE{\(\pi(s|\theta) \leftarrow \pi(s|\theta')\)}
\STATE{\(t \leftarrow t+1\)}
\UNTIL{done}
\end{algorithmic}
\end{algorithm}

We acquire the next trajectory \(\tau_{t+1}\) while concurrently computing the updates for the policy and value networks based on the previously acquired trajectory \(\tau_t\). We first verified in simulation that our off-policy adaptation of \textsc{A3C} would indeed be able to successfully learn a policy to solve the marble maze.
If one had access to multiple robots, the robots could act as parallel agents similar to the case of simulation. However, due to practical limitations, we only have access to a single robot and are thus limited to training with a single agent in the real world case.

\section{Implementation}

We have implemented a simulation of the marble maze using MuJoCo~\cite{todorov:2014:mujoco} to simulate the dynamics, and Ogre 3D~\cite{ogre3d:2018} for the appearance. We carefully measured the maze and marble dimensions to accurately reconstruct its 3D geometry. In order to match the simulated dynamics to the real world dynamics, we have tuned the MuJoCO parameters, with static friction, dynamic friction, and damping parameters in particular. For tuning, the maze was inclined to a known orientation, and the marble was released from various pre-determined locations within the maze. Using the markers (see Figure~\ref{fig:maze}) we aligned the images of the simulated maze to the real maze by computing a homography warp. We then empirically tuned the parameters to match the marble oscillations between the simulated and real maze. Learning the parameters instead would be preferable, but this is left as future work. The simulator is executed as a separate process, and communication between controller and simulator is performed via sockets. The simulator receives an action to perform, and returns an image of the updated marble positions and maze orientation, along with a reward (according to Eq.~\ref{reward-func}) and terminal flag.

The policy network consists of two convolutional layers, followed by a fully-connected layer. The input to the network is an 84\(\times\)84 image. A one-hot action vector and the reward are appended to the 256-dim. output of the fully-connected layer and serves as input to an LSTM layer. This part of the network is shared between the policy (actor) and value (critic) network. For the policy network a fully-connected layer with softmax activation computes a distribution over the actions. For the value network, a fully connected layer outputs a single value. We empirically chose \(L_e=3000\) and \(L_{se}=200\).

The (\(s_t, a_t, r_t\))-tuples are stored in a FIFO experience buffer (of length 3000). We keep track of which tuples have zero and non-zero rewards for importance sampling. For reward prediction we (importance) sample three consecutive frames from the experience buffer. The two convolutional layers and fully connected layer are shared from the policy and value networks. Two more fully connected layers determine a distribution over negative, zero or positive rewards.

For pixel change, we compute the average pixel-change for a 20\(\times\)20 grid, for the central 80\(\times\)80 portion of consecutive images. The pixel-change network re-uses the layers up to and including the LSTM layer for the policy and value network. A fully connected layer together with a deconvolution layers predict 20\(\times\)20 pixel change images. At most \(L_e+1\) frames are sampled from the experience buffer, and we compute the L2 loss between the pixel change predicted by the network, and the recorded pixel change over the sampled sequence. Both losses are added to the \textsc{A3C} loss.

The physics parameters are uniformly sampled from a range around the empirically estimated parameter values. Due to the lack of intuitive interpretation of some of the physics parameters, the range was determined by visually inspecting the resulting dynamics to ensure that the dynamics had sufficient variety, but did not lead to instability in the simulation.

For the real setup, the ROS framework is used to integrate the learning with camera acquisition and robot control. The camera is an Intel RealSense R200 and the robot arm is a Mitsubishi Electric Melfa RV-6SL (see Figure~\ref{fig:maze}--Middle). The execution time of a \(1^{\circ}\) rotation command for the robot arm is about 190ms. Forward passes through the networks and additional computation time add up to about 20 or 30ms. Although we can overlap computation and robot command execution to some degree, observations are acquired at a framerate of 4.3Hz, i.e. 233ms intervals, to ensure robot commands are completed entirely before the new state is obtained. We observed that during concurrent network parameter updates the computation time for a forward pass through the policy network increases drastically. If we expect that the robot action cannot be completed before the new state is observed by the camera, we set the action to \textit{no-op} (Sec.~\ref{subsec:tasksetup}). We implemented a simple marble detector to determine when a marble has passed through a gate, in order to provide a reward signal. For learning in simulation we use the same 4.3Hz framerate. Each incremental rotation action is performed over the course of the allotted time interval of 233ms, such that the next state provided by the simulator reflects the situation after a complete incremental rotation.

\section{Results}

\begin{table*}[h]
  \renewcommand{\arraystretch}{1.3}
  \caption{Comparison of online, offline and online fine-tuning steps for TL for a single marble. A robustified policy can reduce the training steps by a factor of almost 60\(\times\) compared to online training, and a factor of more than 3\(\times\) compared to non-robustified TL fine-tuning.}\label{tbl:compare}
  \centering
  \begin{tabular}{ | l | c | c | c |}
    \hline
    & \textbf{Online (real)} & \textbf{Offline (simulator)} & \textbf{TL (online part)}\\
    \hline
    \textbf{Robust} & \(\sim\)3.5M & \(\sim\)4.0M & \(\sim\)55K \\
    \hline \hline
    \textbf{Non-Robust} & \(\sim\)3.5M & \(\sim\)4.5M & \(\sim\)220K \\
    \hline
  \end{tabular}
\end{table*}

Table~\ref{tbl:compare} compares the number of steps for training a policy to successfully play a one marble maze game. Training directly on the real robot takes about 3.5M steps. For TL, we compare the number of fine-tuning steps necessary for a robustified policy versus a non-robustified policy (fixed parameters). Training a robustified policy in simulation takes about 4.0M steps, whereas a non-robustified policy takes approximately 4.5M to achieve 100\% success rate. TL of a robustified policy requires about 55K steps to "converge". This is a reduction of nearly 60\(\times\) compared to online training. A non-robustified policy requires at least 3\(\times\) the number of fine-tuning steps in order to achieve the same level of success in solving the maze game.

\begin{figure*}[h]
  \centering
  \includegraphics[width=\textwidth]{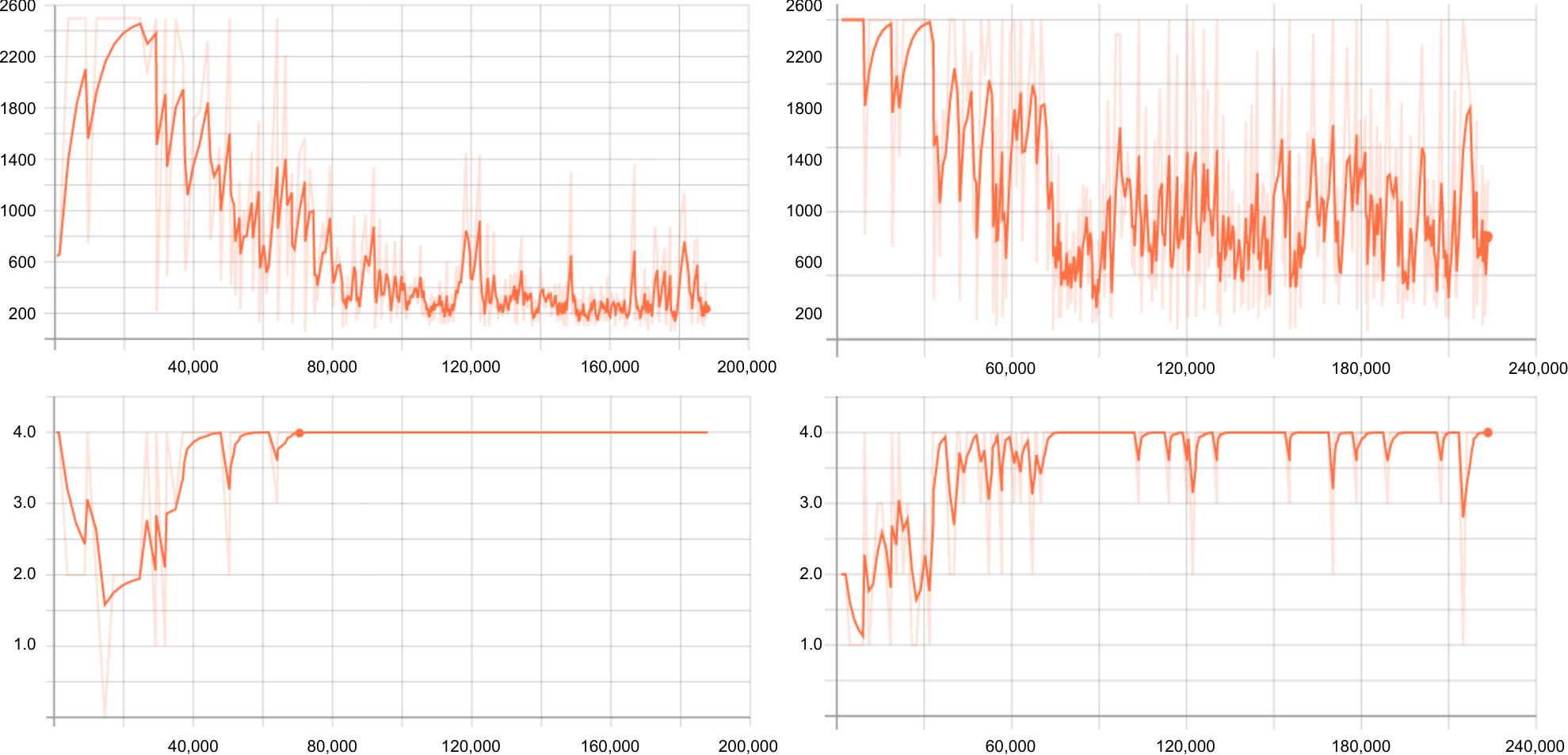}
  \caption{Results for the fine-tuning of policies solving a maze game with one marble for a simulation pre-trained robustified policy (\textbf{Left}), and for a simulation pre-trained non-robustified policy (\textbf{Right}). Note that the horizontal axis ranges between Right and Left are slightly different. In the \textbf{Top} row we plot the number of steps per episode \textemdash with maximum episode length \(L_e=2500\) \textemdash and in the \textbf{Bottom} row we plot the accumulated rewards per episode. The fine-tuning of the robustified policy leads to earlier success of consistently solving the maze game. We consider convergence at \textasciitilde 55K for the robustified policy. Even after more than \textasciitilde 220K fine-tuning episodes, the non-robustified policy occasionally fails to solve the maze game. In addition, the number of steps on average per episode to solve the maze game is significantly less for the case of the robustified policy.}\label{fig:comparison}
\end{figure*}

Figure~\ref{fig:comparison} further shows the benefit for TL of a robustified policy. The left side of Figure~\ref{fig:comparison} shows results for the robustified policy, with results for the non-robustified policy on the right. The bottom row shows the accumulated rewards for an episode. An accumulated reward of 4.0 means that the marble has been maneuvered from the outside ring into the center, since there are four gates to pass through. The graph for the robustified policy shows that the learning essentially converges, i.e., achieve 100\% success, whereas for the non-robustified policy transfer, the success rate is around 90\%. The top row of Figure~\ref{fig:comparison} shows the length of each episode. It is evident that the robustified policy has successfully learned how to handle the complex dynamics to solve the maze game.

We repeated the same experiment for a two marble maze game, with the goal to get both marbles into the center of the maze. We only compared TL with the robustified policy. The results are shown in Table~\ref{tbl:comparetwo}. Learning a two marble game in simulation with \(\pm1\) rewards achieved 100\% success. However, training on the real setup with these rewards proved very challenging. We believe this is due to the geometry of the maze---the center has only one gate, surrounded by four gates in the adjacent ring---coupled with the static friction. We designed a reward function which gives more importance for passing through gates into \textit{rings} closer to the goal. This promotes a marble to stay in the center area, while the controller maneuvers the remaining marble. The rewards were modified to \(\pm\{1, 2, 4, 8\}\) instead (which was also used for training the two marble game offline). When learning online, even after 1M steps, the success rate is still at 0\% (a single marble reached the center about a dozen of times). With fine-tuning a transferred robustified policy, after \(\sim\)225K steps around a 75\% success rate is achieved.

\begin{table}[h]
  \renewcommand{\arraystretch}{1.3}
  \caption{Comparing TL for a two marble maze game. Both the number of steps and success rate are reported.}\label{tbl:comparetwo}
  \centering
  \begin{tabular}{ | l | c | c | c |}
    \hline
    & \textbf{Online} & \textbf{Offline} & \textbf{TL}\\
    \hline
    \textbf{Robust} & \(\sim\)1M (0\%) & \(\sim\)3.0M (100\%) & \(\sim\)225K (75\%) \\
    \hline
  \end{tabular}
\end{table}

We investigate if the transfer of a single marble policy learned offline, would require longer fine-tuning for a two marble game online. After \(\sim\)100K steps of fine-tuning, the policy was able to start solving the game. A success rate of about 50\% was achieved after \(\sim\)400K steps. Thus, fine-tuning a robustified policy trained on a two marble maze game in simulation achieves a higher success rate compared to the fine-tuning of a single marble robustified policy.

We refer the reader to the supplemental material for videos of example roll-outs for single and two marble maze games.

\section{Discussion and Future Work}

Deep reinforcement learning is capable of learning complicated robot tasks, and in some cases achieving (beyond) human-level performance. Deep RL requires many training samples, especially in the case of model-free approaches. For learning robot tasks, learning in simulation is desirable since robots are slow, can be dangerous and are expensive. Powerful GPUs and CPUs have enabled simulation of complex dynamics coupled with high quality rendering at high speeds. Transfer learning, i.e., the training in simulation and subsequent transfer to the real world, is typically followed by fine-tuning. Fine-tuning is necessary to adapt to any differences between the simulated and the real world. Previous work has focused on transfer learning tasks involving linear dynamics, such as controlling a robot to pick an object and place it at some desired location. However, we explore the case when the dynamics are complex. Non-linearities arise due to static and dynamic friction, acceleration and collisions of objects interacting with each other and the environment. We compare learning online, i.e., directly in the real world, with learning in simulation where the physics, appearance and system parameters are varied during training. For reinforcement learning we refer to this as learning robustified policies. We show that the time required for fine-tuning with robustified policies, is greatly reduced.

Although we have shown that model-free deep reinforcement learning can be successfully used to learn tasks involving complex dynamics, there are drawbacks of using a model-free approach. In the example discussed in our paper, the dynamics are (mostly) captured by the LSTM layer in the network. In the case of more than one marble the amount of fine-tuning time significantly increases. In general, as the complexity of the state space increases, the amount of training time increases as well. When people perform tasks such as the maze game, they typically have a decent prediction of where the marble(s) will go given the amount of rotation applied. In~\cite{wu:2015:galileo, wu:2017:deanimation} the graphics and physics engine are embedded within the learning to recover physics parameters and perform predictions of the dynamics. In~\cite{ehrhardt:2018:intuphysics} the physics and dynamics predictions are modeled with networks. These approaches are interesting research directions for tasks involving complex dynamics.

We currently use high-dimensional images as input to the learning framework. Low-dimensional input, i.e. marble position and velocity, may be used instead. In addition, rather than producing a distribution over a discrete set of actions, the problem can be formulated as a regression instead and directly produce values for the \(x\) and \(y\) axes rotations~\cite{lillicrap:2015:contcontrol, mnih:2016:A3C}.

People quickly figure out that the task can be broken down into moving a single marble at the time into the center, while avoiding marbles already in the center location from spilling back out. Discovering such sub-tasks automatically would be another interesting research direction. Along those lines, teaching a robot to perform tasks by human demonstration, or imitation learning, could teach robots complicated tasks without the need for elaborate reward functions, e.g.,~\cite{finn:2017:oneshotimitation}.

\section*{ACKNOWLEDGEMENTS}

We want to thank Rachana Sreedhar for the implementation of the simulator and Wei-An Lin for the Pytorch implementation of deep reinforcement learning.

\bibliographystyle{IEEEtran}
\bibliography{bibliography}

\end{document}